\documentclass[11pt, a4paper, logo, twocolumn, external, copyright]{googledeepmind}

\pdftrailerid{redacted}

\makeatletter
\renewcommand\bibentry[1]{\nocite{#1}{\frenchspacing\@nameuse{BR@r@#1\@extra@b@citeb}}}
\makeatother
\usepackage{kantlipsum, lipsum}
\usepackage{dsfont}
\usepackage{gdm-colors}
\usepackage{cleveref}
\usepackage[absolute,overlay]{textpos}

\usepackage{fancyhdr}
\usepackage{bm}

\usepackage[authoryear, sort&compress, round]{natbib}

\graphicspath{{figures/}}

\newcommand{\genieemoji}{\includegraphics[height=1.3\fontcharht\font`A]{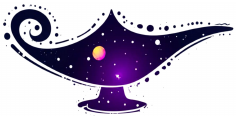}}

\newcommand{\x}{\bm{x}}
\newcommand{\z}{\bm{z}}
\renewcommand{\a}{\bm{a}}

\newcommand{\psnrdiff}{\ensuremath{\Delta_t\text{PSNR}}}

\title{\genieemoji{} Genie: Generative Interactive Environments}

\correspondingauthor{Ashley Edwards (\href{mailto:edwardsashley@google.com}{edwardsashley@google.com}), Jack Parker-Holder (\href{mailto:jparkerholder@google.com}{jparkerholder@google.com}).}


\author[*,1]{Jake Bruce}
\author[*,1]{Michael Dennis}
\author[*,1]{Ashley Edwards}
\author[*,1]{Jack Parker-Holder}
\author[*,1]{Yuge (Jimmy) Shi}
\author[1]{Edward Hughes}
\author[1]{Matthew Lai}
\author[1]{Aditi Mavalankar}
\author[1]{Richie Steigerwald}
\author[1]{Chris Apps}
\author[1]{Yusuf Aytar}
\author[1]{Sarah Bechtle}
\author[1]{Feryal Behbahani}
\author[1]{Stephanie Chan}
\author[1]{Nicolas Heess}
\author[1]{Lucy Gonzalez}
\author[1]{Simon Osindero}
\author[1]{Sherjil Ozair}
\author[1]{Scott Reed}
\author[1]{Jingwei Zhang}
\author[1]{Konrad Zolna}
\author[1,2]{Jeff Clune}
\author[1]{Nando de Freitas}
\author[1]{Satinder Singh}
\author[*,1]{Tim Rocktäschel}

\affil[*]{Equal contributions}
\affil[1]{Google DeepMind}
\affil[2]{University of British Columbia}

\banner{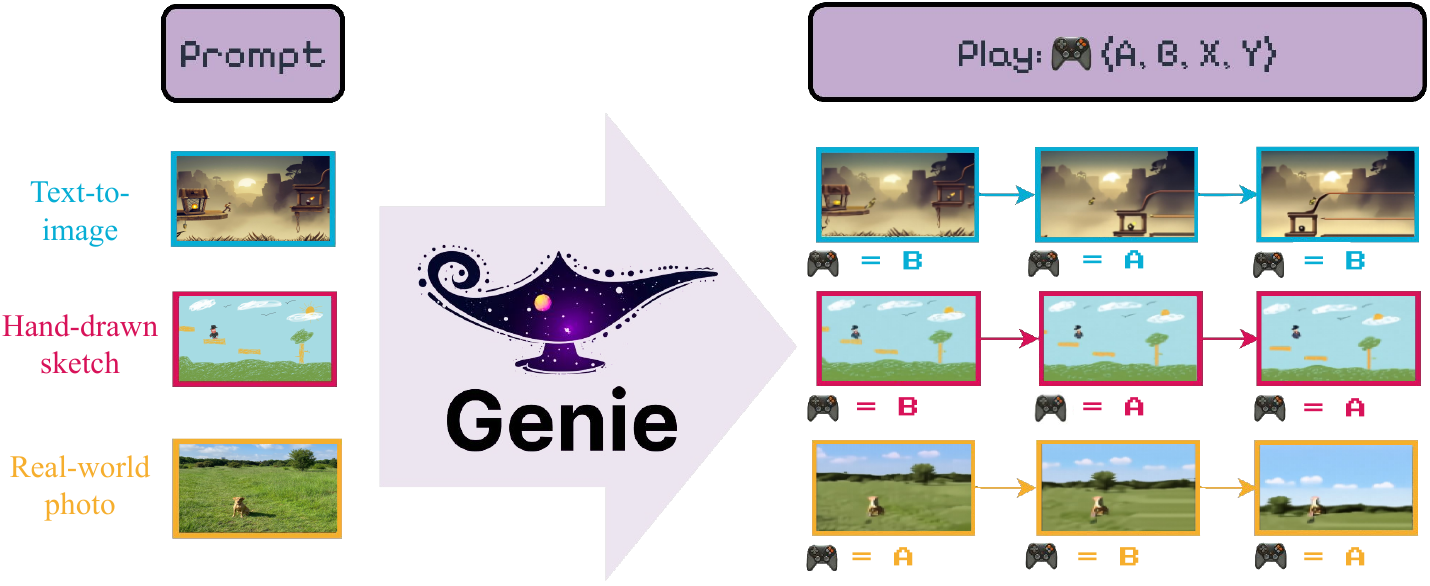}{\textbf{A whole new world}: Genie is capable of converting a variety of different prompts into interactive, playable environments that can be easily created, stepped into, and explored. This is made possible via a latent action interface, learned fully unsupervised from Internet videos. On the right we see a few generated steps for taking two latent actions. See more examples on our \href{https://sites.google.com/view/genie-2024/home}{website}.}

\begin{abstract}
We introduce Genie, the first \emph{generative interactive environment} trained in an unsupervised manner from unlabelled Internet videos. The model can be prompted to generate an endless variety of action-controllable virtual worlds described through text, synthetic images, photographs, and even sketches. At 11B parameters, Genie can be considered a \emph{foundation world model}. It is comprised of a spatiotemporal video tokenizer, an autoregressive dynamics model, and a simple and scalable latent action model. Genie enables users to act in the generated environments on a frame-by-frame basis \emph{despite training without any ground-truth action labels} or other domain-specific requirements typically found in the world model literature. Further the resulting learned latent action space facilitates training agents to imitate behaviors from unseen videos, opening the path for training generalist agents of the future.
\end{abstract}

\begin{document}
\maketitle
\newpage
\clearpage
\section{1. Introduction}

The last few years have seen an emergence of \emph{generative AI}, with models capable of generating novel and creative content. Driven by breakthroughs in architectures such as transformers \citep{vaswani2017attention}, advances in hardware, and a recent focus on scaling models and datasets, we can now generate coherent, conversational language \citep{radford2018improving, radford2019language, brown2020language}, as well as crisp and aesthetically pleasing images from a text prompt \citep{ramesh21a, ramesh2022hierarchical, saharia2022photorealistic, Rombach_2022_CVPR}. Early signs indicate video generation will be yet another frontier, with recent results suggesting that such models may also benefit from scale \citep{hong2023cogvideo, ho2022imagen, esser2023structure, blattmann2023stable}. Still, there remains a gulf between the level of interactions and engagement of video generative models and language tools such as ChatGPT, let alone more immersive experiences.

\begin{figure}[htb]
\centering
\includegraphics[width=0.99\linewidth]{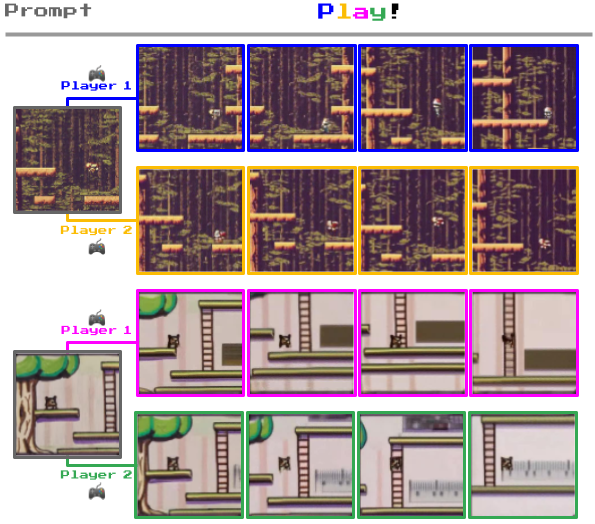}
\caption{\textbf{Diverse trajectories}: Genie is a generative model that can be used as an interactive environment. The model can be prompted in various ways, either with a generated image (top) or a hand-drawn sketch (bottom). At each time step, the model takes a user-provided latent action to generate the next frame, producing trajectories with interesting and diverse character actions. }\label{fig:platformer_trajectories}
\end{figure}

What if, given a large corpus of videos from the Internet, we could not only train models capable of generating novel images or videos, but entire interactive experiences? We propose \emph{generative interactive environments}, a new paradigm for generative AI whereby interactive environments can be generated from a single text or image prompt. Our approach, Genie, is trained from a large dataset of over 200,000 hours of publicly available Internet gaming videos and, despite training \emph{without action or text annotations}, is controllable on a frame-by-frame basis via a learned latent action space (see \Cref{tab:genie_related} for a comparison to other approaches). At 11B parameters, Genie exhibits properties typically seen in foundation models---it can take an unseen image as a prompt making it possible to create and play entirely imagined virtual worlds (e.g \Cref{fig:platformer_trajectories}).

Genie builds on ideas from state-of-the-art video generation models \citep{villegas2023phenaki, gupta2023maskvit}, with a core design choice being spatiotemporal (ST) transformers \citep{xu2020spatial} which are used in all of our model components. Genie utilizes a novel video tokenizer, and extracts latent actions via a causal action model. Both the video tokens and latent actions are passed to a dynamics model, which autoregressively predicts the next frame using MaskGIT \citep{Chang_2022_CVPR}. We provide a rigorous scaling analysis of our architecture with respect to both batch and model size, which we vary from 40M to 2.7B parameters. The results show that our architecture scales gracefully with additional computational resources, leading to a final 11B parameter model. We train Genie on a filtered set of 30,000 hours of Internet gameplay videos from hundreds of 2D platformer games, producing a foundation world model for this setting. 

To demonstrate the generality of our approach, we also train a separate model on action-free robot videos from the RT1 dataset \citep{brohan2023rt1}, learning a generative environment with consistent latent actions. Finally, we show that latent actions learned from Internet videos can be used for inferring policies from unseen action-free videos of simulated reinforcement learning (RL) environments, indicating that Genie may hold the key to unlocking unlimited data for training the next generation of generalist agents \citep{xland, ada, reed2022a, clune2019ai}.

\begin{figure*}[h]
    \centering
    \vspace{-3mm}
    \includegraphics[width=0.99\linewidth]{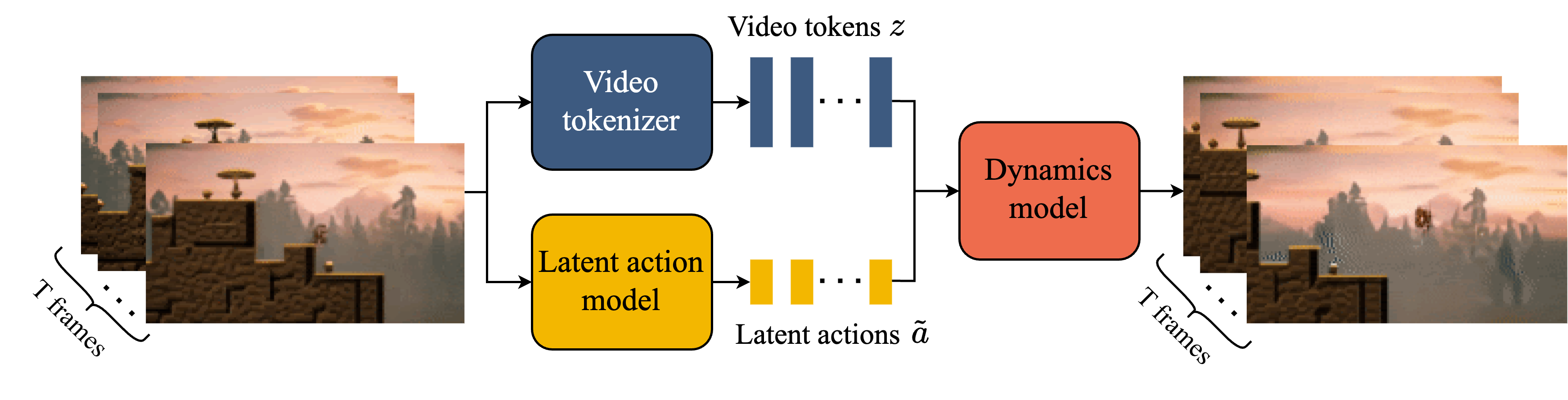}
    \vspace{-2mm}
    \caption{\textbf{Genie model training}:  Genie takes in $T$ frames of video as input, tokenizes them into discrete tokens $\z$ via the video tokenizer, and infers the latent actions $\tilde{\a}$ between each frame with the latent action model. Both are then passed to the dynamics model to generate predictions for the next frames in an iterative manner.}
    \vspace{-2mm}
    \label{fig:genie_architecture}
\end{figure*}

\begin{table}[h]
\caption{\textbf{A new class of generative model}: Genie is a novel video and world model that is controllable on a frame-by-frame basis, which requires \textbf{only video data} at train time.}
\label{tab:genie_related}
\centering
\resizebox{.48\textwidth}{!}{
\begin{tabular}{llc}
\toprule
Model Class  & Training Data & Controllability  \\ \midrule
World Models  & Video + Actions & Frame-level \\ 
Video Models  & Video + Text & Video-level \\
\rowcolor{blue!60!red!15}
Genie  & Video & Frame-level \\ \bottomrule
\end{tabular}
}
\end{table}

\section{2. Methodology}
Genie is a generative interactive environment trained from video-only data. In this section we begin with preliminaries before explaining the main components of our model.

\begin{figure}[h]
\centering
\includegraphics[width=0.99\linewidth]{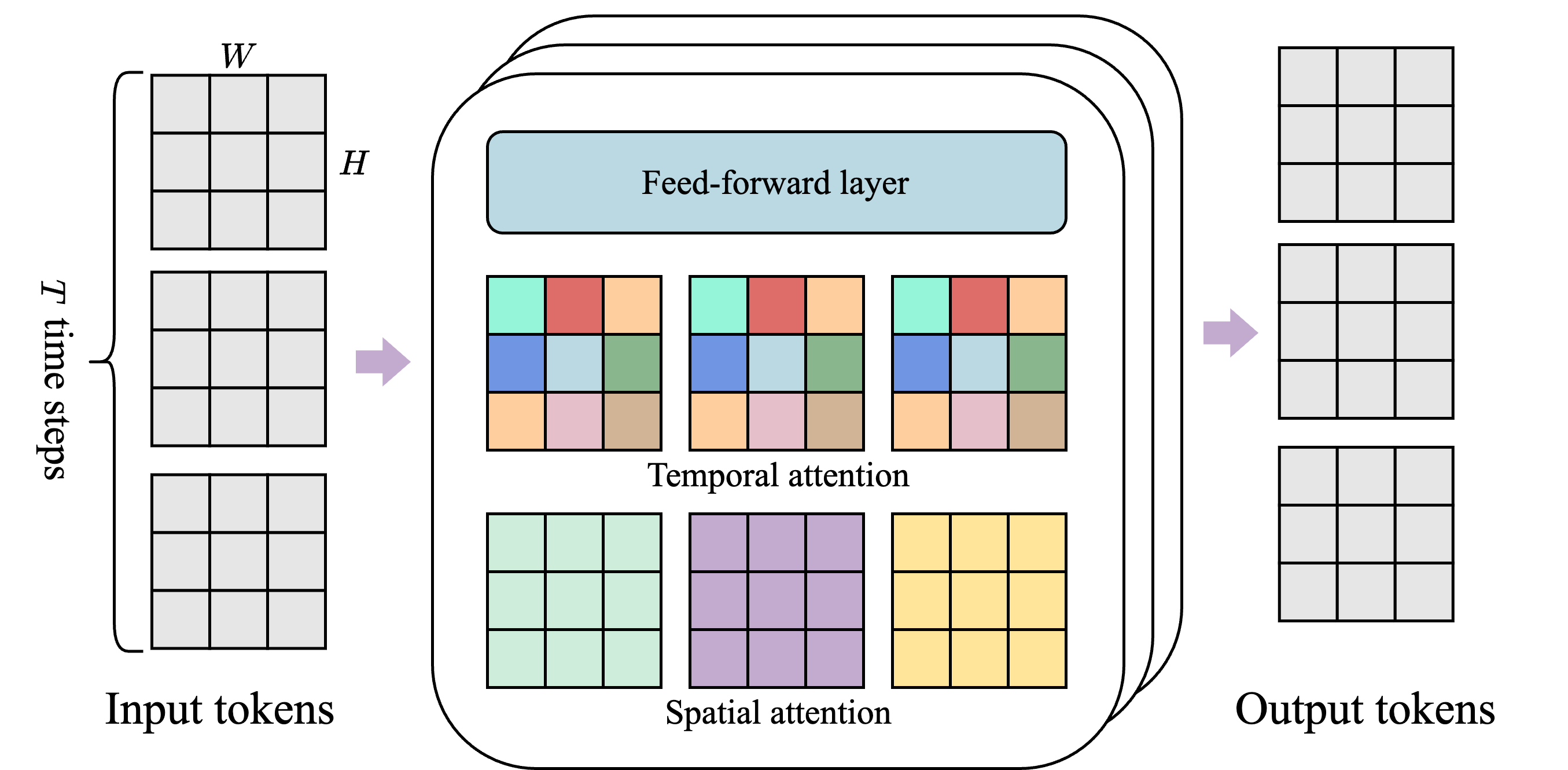}
\caption{\textbf{ST-transformer architecture}. The architecture is composed of $L$ spatiotemporal blocks, each containing a spatial layer, temporal layer and feed-forward layer. Each color represents a single self-attention map, with the spatial layer attending over the $H\times W$ tokens from within a single time step, and temporal the same token from across the $T$ time steps.}\label{fig:st_architecture}
\end{figure}

Several components in the Genie architecture are based on the Vision Transformer (ViT) \citep{vaswani2017attention, dosovitskiy2021an}. Notably, the quadratic memory cost of transformers poses challenges for videos, which can contain up to $O(10^4)$ tokens. We thus adopt a memory efficient  ST-transformer architecture (inspired by \citet{xu2020spatial}, see \Cref{fig:st_architecture}) across all model components, balancing model capacity with computational constraints.

Unlike a traditional transformer where every token attends to all others, an ST-transformer contains $L$ spatiotemporal blocks with interleaved spatial and temporal attention layers, followed by a feed-forward layer (FFW) as standard attention blocks. The self-attention in the spatial layer attends over the $1\times H\times W$ tokens within each time step, and in the temporal layer attends over $T\times 1 \times 1$ tokens across the $T$ time steps. Similar to sequence transformers, the temporal layer assumes a causal structure with a causal mask. Crucially, the dominating factor of computation complexity (i.e. the spatial attention layer) in our architecture scales linearly with the number of frames rather than quadratically, making it much more efficient for video generation with consistent dynamics over extended interactions. Further, note that in the ST block, we include only one FFW after both spatial and temporal components, omitting the post-spatial FFW to allow for scaling up other components of the model, which we observe to improve results significantly.

\subsection{Model Components}
\label{sec:model_components}

As shown in \Cref{fig:genie_architecture}, our model contains three key components: 1) a \emph{\textbf{latent action model}} that infers the latent action $\a$ between each pair of frames and 2) a \emph{\textbf{video tokenizer}} that converts raw video frames into discrete tokens $\z$ and 3) a \emph{\textbf{dynamics model}} that, given a latent action and past frame tokens, predicts the next frame of the video. The model is trained in two phases following a standard autoregressive video generation pipeline: we train the video tokenizer first, which is used for the dynamics model. We then co-train the latent action model (directly from pixels) and the dynamics model (on video tokens).

\begin{figure}[h]
\centering
\vspace{-10pt}
\includegraphics[width=0.9\linewidth]{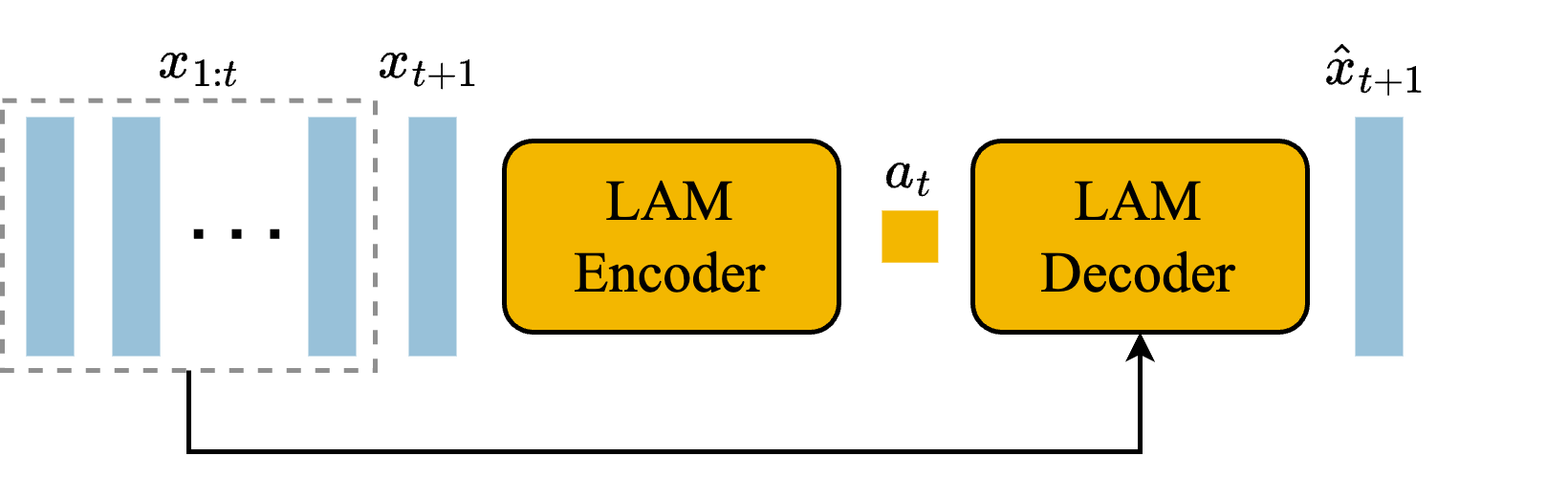}
\vspace{-5pt}
\caption{\textbf{Latent action model}: learns actions $a_t$ unsupervised from unlabelled video frames.}\label{fig:lam_architecture}
\vspace{-5pt}
\end{figure}

\textbf{Latent Action Model (LAM)} To achieve controllable video generation, we condition each future frame prediction on the action taken at the previous frame. However, such action labels are rarely available in videos from the Internet and action annotation can be costly to obtain. Instead, we learn \emph{latent actions} in a fully unsupervised manner (see \Cref{fig:lam_architecture}).

First, an encoder takes as inputs all previous frames $\x_{1:t}=(x_1,\cdots x_t)$ as well as the next frame $x_{t+1}$, and outputs a corresponding set of continuous latent actions $\tilde{\a}_{1:t}=(\tilde{a}_1,\cdots \tilde{a}_t)$. A decoder then takes all previous frames and latent actions as input and predicts the next frame $\hat{x}_{t+1}$.

To train the model, we leverage a VQ-VAE-based objective \citep{vqvae2017}, which enables us to limit the number of predicted actions to a small discrete set of codes. We limit the vocabulary size $|A|$ of the VQ codebook, i.e. the maximum number of possible latent actions, to a small value to permit human playability and further enforce controllability (we use $|A|=8$ in our experiments). As the decoder only has access to the history and latent action, $\tilde{a}_t$ should encode the most meaningful changes between the past and the future for the decoder to successfully reconstruct the future frame. Note that this decoder exists only to give the LAM training signal. In fact, apart from the VQ codebook, the entire LAM is discarded at inference time and replaced with actions from the user.

We utilize our ST-transformer architecture for the latent action model. The causal mask in the temporal layer allows us to take the entire video $\x_{1:T}$ as input and generate all latent actions between each frame $\tilde{\a}_{1:T-1}$.

\begin{figure}[h!]
\centering
\includegraphics[width=0.9\linewidth]{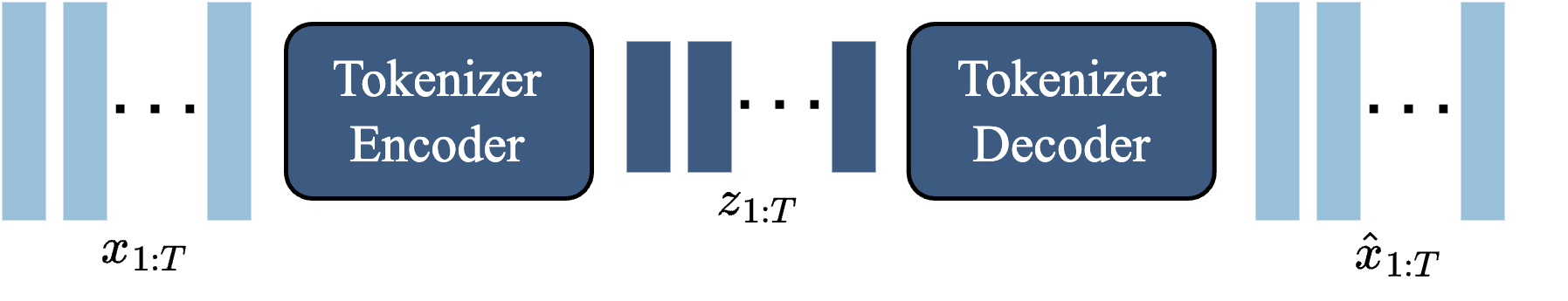}
\caption{\textbf{Video tokenizer}: a VQ-VAE with ST-transformer.}\label{fig:tokenizer_architecture}
\end{figure}

\textbf{Video Tokenizer} Following prior work \citep{villegas2023phenaki, gupta2023maskvit, teco}, we compress videos into discrete tokens to reduce dimensionality and enable higher quality video generation (see \Cref{fig:tokenizer_architecture}). We again make use of VQ-VAE, which takes in $T$ frames of video $\x_{1:T}=(x_1, x_2,\cdots,x_T) \in \mathbb{R}^{T \times H \times W \times C}$ as input, generating discrete representations for each frame  $\z_{1:T}=(z_1, z_2,\cdots,z_T) \in \mathbb{I}^{T\times D}$, where $D$ is the size of the discrete latent space. The tokenizer is trained using a standard VQ-VQAE objective over the entire video sequence.

Unlike prior works that focus on spatial-only compression in the tokenization phase \citep{hong2022cogvideo,wu2022nuwa, gupta2023maskvit}, we utilize the ST-transformer in both the encoder and decoder to incorporate temporal dynamics in the encodings, which improves the video generation quality. By the causal nature of the ST-transformer, each discrete encoding $z_t$ contains information from all previously seen frames of the video $\x_{1:t}$. Phenaki \citep{villegas2023phenaki} also uses a temporal-aware tokenizer, C-ViViT, but this architecture is compute intensive, as the cost grows quadratically with the number of frames---in comparison, our ST-transformer based tokenizer (ST-ViViT) is much more compute efficient with the dominating factor in its cost increasing linearly with the number of frames.

\begin{figure}[h!]
\centering
\includegraphics[width=0.7\linewidth]{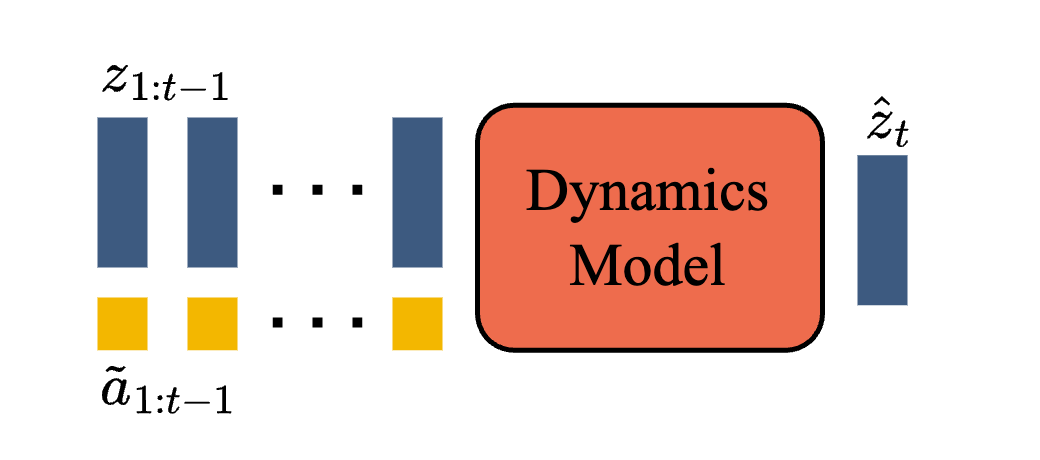}
\caption{\textbf{Dynamics model}: takes in video tokens and action embeddings, and predicts future masked video tokens.}\label{fig:dynamics_architecture}
\end{figure}

\textbf{Dynamics Model} The dynamics model is a decoder-only MaskGIT \citep{Chang_2022_CVPR} transformer (\Cref{fig:dynamics_architecture}). At each time step $t\in [1, T]$, it takes in the tokenized video $\z_{1:t-1}$ and stopgrad latent actions $\tilde{\a}_{1:t-1}$ and predicts the next frame tokens $\hat{z}_{t}$.
We again utilize an ST-transformer, whose causal structure enables us to use tokens from all $(T-1)$ frames $\z_{1:T-1}$ and latent actions $\tilde{\a}_{1:T-1}$ as input, and generate predictions for all next frames $\hat{\z}_{2:T}$. The model is trained with a cross-entropy loss between the predicted tokens $\hat{\z}_{2:T}$ and ground-truth tokens $\z_{2:T}$. At train time we randomly mask the input tokens $\z_{2:T-1}$ according to a Bernoulli distribution masking rate sampled uniformly between $0.5$ and $1$.
Note that a common practice for training world-models, including transformer-based models, is to concatenate the action at time $t$ to the corresponding frame \citep{micheli2023transformers, robine2023transformerbased}. However, we found that treating the latent actions as~\emph{additive embeddings} for both the latent action and dynamics models helped to improve the controllability of the generations.

\subsection{Inference: Action-Controllable Video Generation}

\begin{figure}[h!]
    \centering
    \includegraphics[width=0.99\linewidth]{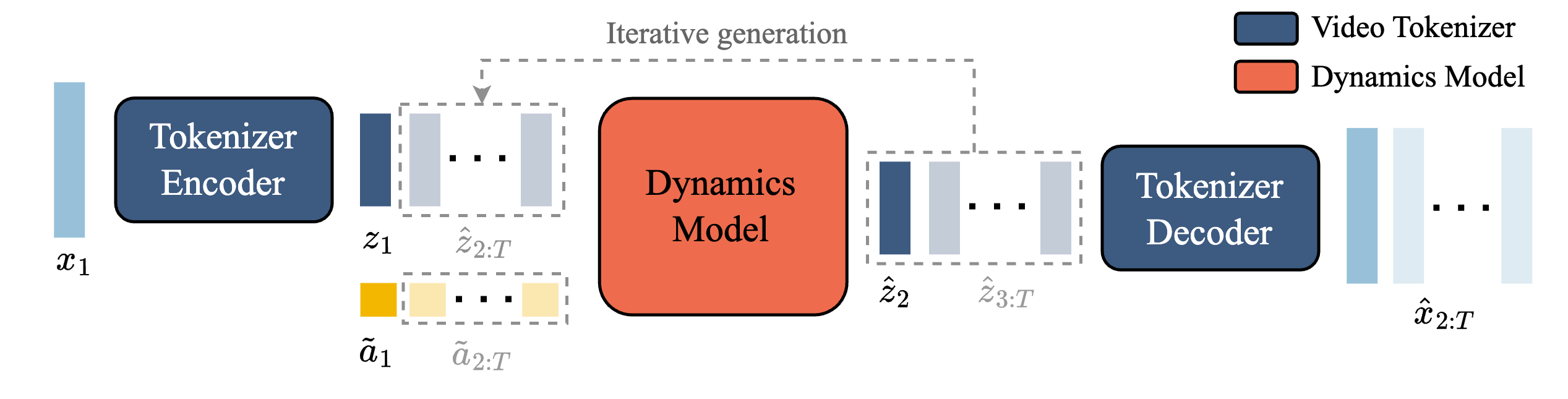}
    \caption{\textbf{Genie Inference}: the prompt frame is tokenized, combined with the latent action taken by the user, and passed to the dynamics model for iterative generation. The predicted frame tokens are then decoded back to image space via the tokenizer's decoder.}
    \label{fig:genie_inference}
\end{figure}

We now describe how to use Genie for action-controllable video generation at inference time (see \Cref{fig:genie_inference}). A player first prompts the model with an image $x_1$ that serves as the initial frame\footnote{The model can be conditioned on a varying number of prompt frames. Here we start from one image as an example.}. The image is tokenized using the video encoder, yielding $z_1$. The player then specifies a discrete latent action $a_1$ to take by choosing any integer value within $[0,|A|)$.\footnote{When first interacting with the model, it is unclear how each latent action will impact the next frame generation. However, we found that the meaning of each action \emph{remained consistent} across different inputs. Hence, interpreting the mapping of latent actions is akin to learning the buttons on a new controller.} The dynamics model takes the frame tokens $z_1$ and corresponding latent action $\tilde{a}_1$, which is obtained by indexing into the VQ codebook with the discrete input $a_1$, to predict the next frame tokens $z_2$. This process is repeated to generate the rest of the sequence $\hat{\z}_{2:T}$ in an autoregressive manner as actions continue to be passed to the model, while tokens are decoded into video frames $\hat{\x}_{2:T}$ with the tokenizer's decoder. Note that we can regenerate ground truth videos from the dataset by passing the model the starting frame and inferred actions from the video, or generate completely new videos (or trajectories) by changing the actions.

\section{3. Experimental Results}

\begin{figure*}[t!]
    \centering
    \begin{minipage}{\textwidth}
    \centering\includegraphics[width=\linewidth]{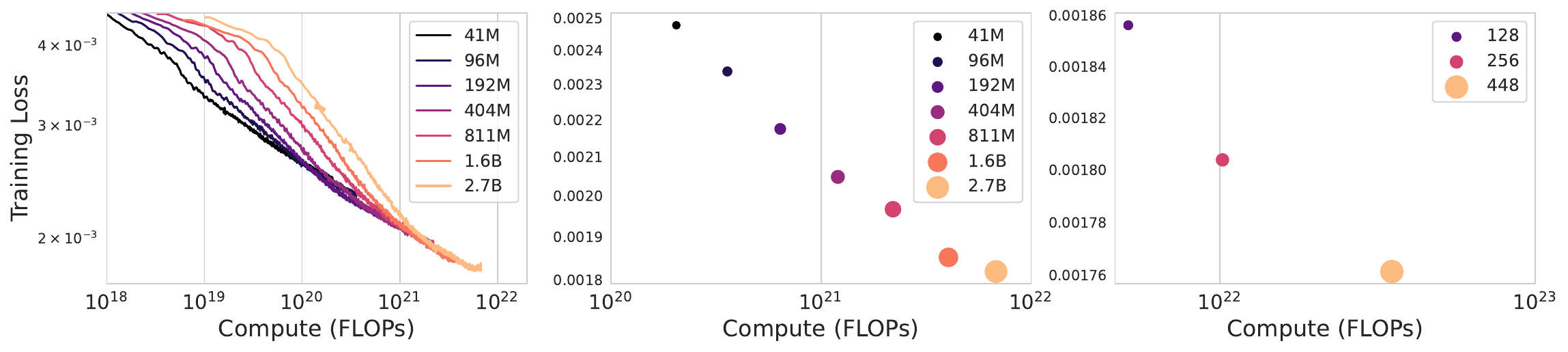}
    \vspace{-4mm}
    \caption{\textbf{Scaling results}. \textbf{Left}: Training curves for different model sizes, \textbf{Middle}: Final training loss for each model size, averaged over the last 300 updates, \textbf{Right}: Final training loss for a 2.3B model with different batch sizes.}
    \vspace{-4mm}
    \label{fig:scaling_params}
    \end{minipage}
\end{figure*}

\textbf{Datasets} We train Genie on a large-scale dataset collected from publicly available Internet videos of 2D Platformer games (referred to from here on as ``Platformers''). We construct the Platformers dataset by filtering publicly available videos for keywords relating to platformers, yielding 55M 16s video clips at 10FPS, with 160x90 resolution. The final dataset contains 6.8M 16s video clips (30k hours), within an order of magnitude of other popular Internet video datasets \citep{wang2023internvid, webvid}. More details can be found in \Cref{appendix:ytdata}. Unless otherwise specified, results are with a 11B-parameter model trained on this dataset. 

To verify the generality of our method, we also consider the robotics datasets used to train RT1 \cite{brohan2023rt1}, combining their dataset of ${\sim}130k$ robot demonstrations with a separate dataset of simulation data and the 209k episodes of real robot data from prior work \citep{kalashnikov2018qt}. Note that we do not use actions from any of these datasets, and simply treat them as videos. For simplicity, from here on we refer to this dataset as ``Robotics''.

\textbf{Metrics} We examine the video generation performance of Genie via two factors, namely \emph{\textbf{video fidelity}}, i.e. the quality of video generation, and \emph{\textbf{controllability}}, i.e. how much impact the latent actions have in video generation. 
For video fidelity we use the Frechet Video Distance (FVD), a video-level metric, which has been shown to have a high level of alignment to human evaluation on video quality \citep{unterthiner2019fvd}. For controllability, we devise a metric based on peak signal-to-noise ratio (PSNR) which we call \psnrdiff, that measures how much the video generations differ when conditioned on latent actions inferred from ground-truth ($\hat{x}_t$) vs. sampled from a random distribution ($\hat{x}_t'$):
\begin{align}
     \psnrdiff = \text{PSNR}(x_t, \hat{x}_t) - \text{PSNR}(x_t, \hat{x}_t'), \notag
\end{align}
where $x_t$ denotes the ground-truth frame at time $t$, $\hat{x}_t$ denotes the frame from latent actions $\tilde{\a}_{1:t}$ inferred from ground-truth frames, and $\hat{x}_t'$ the same frame generated from a sequence of latent actions randomly sampled from a categorical distribution. As such, the greater \psnrdiff\ is, the more the video generated from random latent actions differs from ground-truth, which indicates a higher level of controllability from the latent actions. For all experiments we report \psnrdiff\ with $t=4$.

\textbf{Training Details} Our video tokenizer uses 200M parameters, a patch size of 4 and a codebook with embedding size 32 and 1024 unique codes, which we found to be the most effective given the trade-off between reconstruction quality of the tokenizer and downstream performance of video prediction. The latent action model has 300M parameters, a patch size of 16, and a codebook with embedding size 32 and 8 unique codes (latent actions). For all modelling components we use a sequence length of 16 frames with an FPS of 10. Further, we employ bfloat16 and QK norm for training our dynamics model, which has been shown to stabilize training at large scale \citep{henry-etal-2020-query,pmlr-v202-dehghani23a}. At inference time, we perform 25 MaskGIT steps for the sampling of each frame with a temperature of 2 using random sampling. See \Cref{sec:app_training} for more details. 
\subsection{Scaling Results}
In this section, we investigate the scaling behavior of our model. To this end, we conduct studies that explore the impact of both model size and batch size. See \Cref{sec:app_scaling_details} for more details on architecture and compute usage. 

\begin{figure*}[t!]
    \centering
    \vspace{-3mm}
    \includegraphics[width=0.94\textwidth]{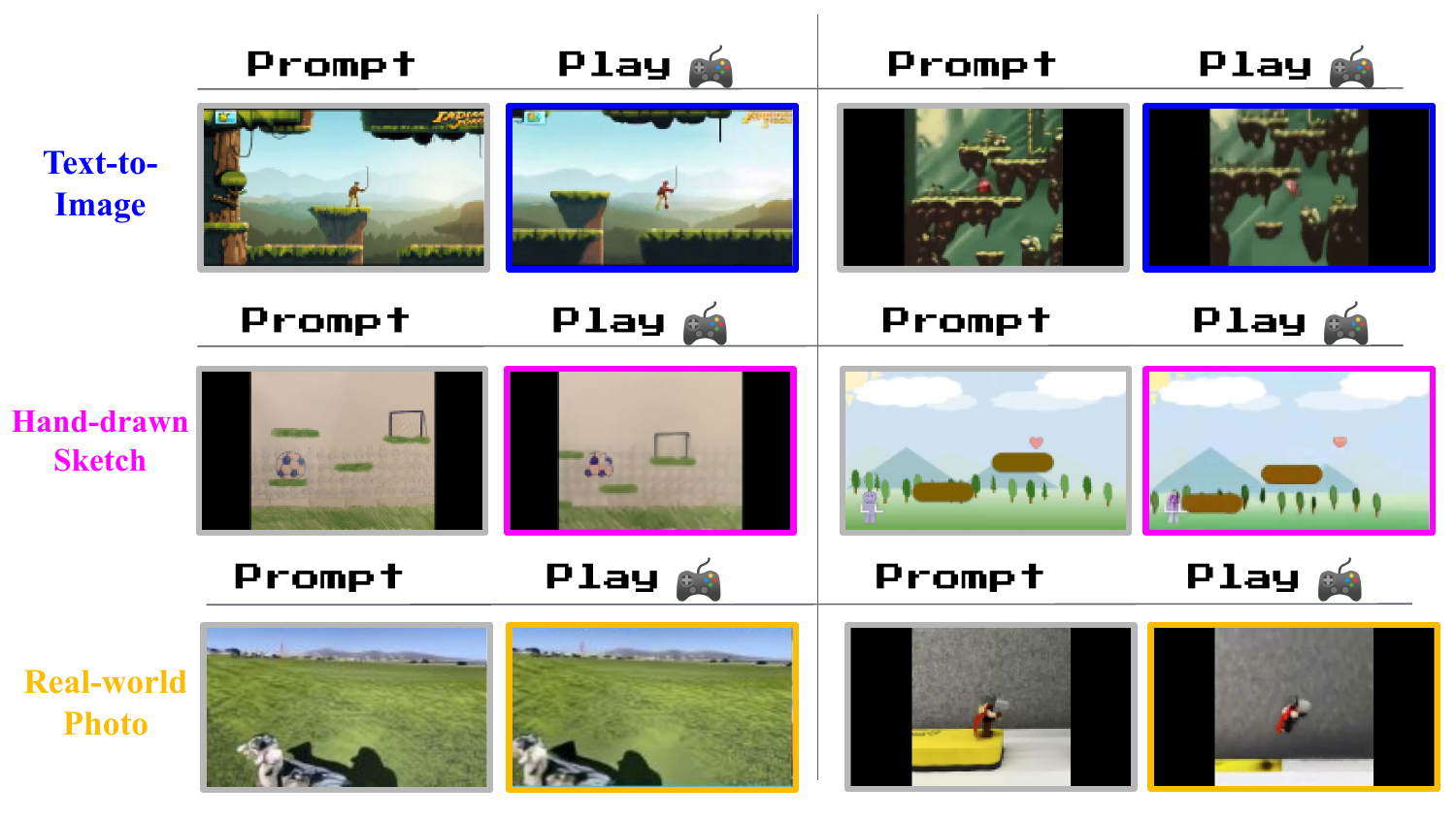}
    \vspace{-2mm}
    \caption{\textbf{Playing from Image Prompts}: We can prompt Genie with images generated by text-to-image models, hand-drawn sketches or real-world photos. In each case we show the prompt frame and a second frame after taking one of the latent actions four consecutive times. In each case we see clear character movement, despite some of the images being visually distinct from the dataset.}
    \label{fig:action_grid_platformers_emergent}
\end{figure*}

\textbf{Scaling Model Size} Given a fixed video tokenizer and action model architecture, we train a series of dynamics models ranging from 40M to 2.7B parameters. \Cref{fig:scaling_params} shows our architecture scales gracefully with model parameters, with each increase in size corresponding to a consistent decrease in the final training loss. This is a strong indication that our approach benefits from scaling, which we exploit with our main Genie model.

\textbf{Scaling Batch Size} We also investigate the effect of scaling the batch size, considering a 2.3B model with batch sizes of 128, 256, and 448, equating to 1.9M, 3.8M and 6.6M tokens. As shown in Figure~\ref{fig:scaling_params}, increasing the batch size leads to a similarly favorable gain in terms of model performance.

\textbf{Genie Model} It is clear that increasing both model size and batch size helps improve model performance. As a result, for our final model, we train a 10.1B dynamics model with a batch size of 512, for a total of 125k steps, using 256 TPUv5p. When combined with the tokenizer and action model this brings the total to 10.7B parameters, trained on 942B tokens, which we refer to as the Genie model. For our website, we train a larger decoder mapping tokens to 360p videos, adding additional parameters.

\subsection{Qualitative Results}

We now present qualitative results from the Genie model. We showcase a 11B parameter model trained on the Platformers dataset and a smaller model trained on the Robotics dataset. Our model generates high-quality, controllable videos across diverse domains. Notably, we qualitatively evaluate our Platformers-trained model using \emph{only out-of-distribution (OOD) image prompts}, including those generated from text-to-image models, hand-drawn sketches, and even realistic photos. The ability to generalize to such significantly OOD inputs underscores the robustness of our approach and the value of training on large-scale data, which would not have been feasible with real actions as input.

\begin{figure*}[t!]
    \centering
    \includegraphics[width=\textwidth]{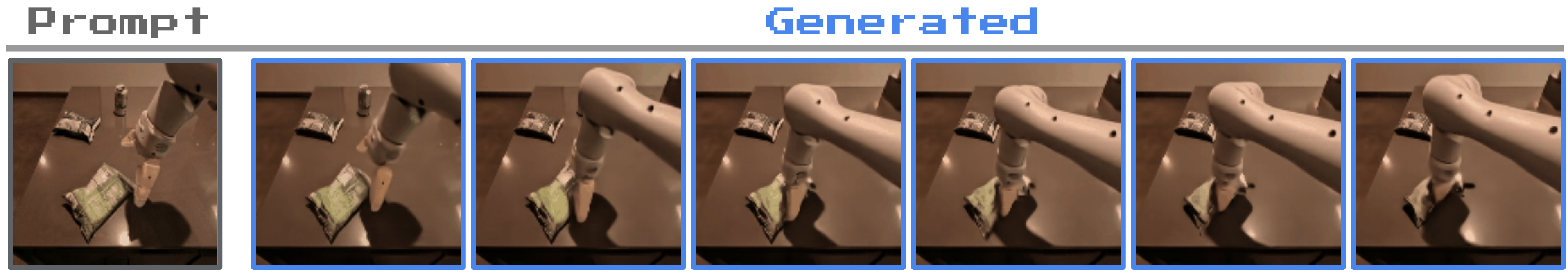}
    \caption{\textbf{Learning to simulate deformable objects}: we show frames from a ten step trajectory in the model, taking the same action. Genie is capable of learning the physical properties of objects such as bags of chips.}
    \label{fig:chips}
\end{figure*}

\textbf{Platformers-trained model} \Cref{fig:action_grid_platformers_emergent} showcases examples of our model's generations prompted from OOD images, including (top row) images generated from Imagen2 \citep{ho2022imagen, imagen2}, (second row) hand-drawn sketches and (bottom row) real-world photos. Genie is able to bring these imagined worlds to life, as we see game-like behaviour when interacting with each example. We showcase more generations by our model in \Cref{appendix:more_example}, additionally highlighting the consistency of the latent actions.

\begin{figure}[h!]
\centering
\includegraphics[width=0.93\linewidth]{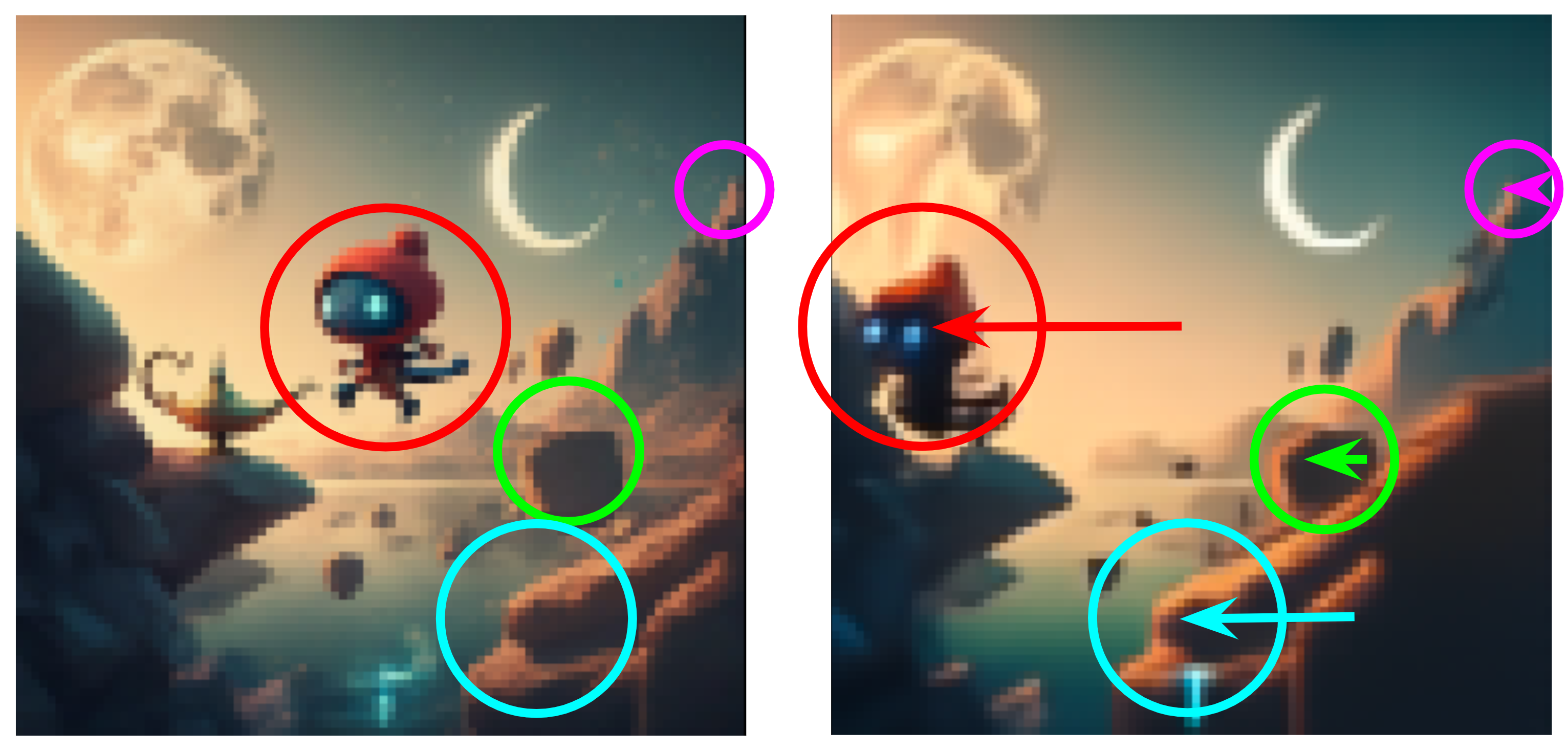}
\vspace{-2mm}
\caption{\textbf{Emulating parallax}, a common feature in platformer games. From this initial text-generated image, the \textcolor{red}{\textbf{foreground}} moves more than the \textcolor{cyan}{\textbf{near}} and \textcolor{green}{\textbf{far}} middle ground, while the \textcolor{violet}{\textbf{background}} moves only slightly.}\label{fig:parallax}
\vspace{-2mm}
\end{figure}

Another emergent capability of our model is its ability to understand 3D scenes and emulate parallax, which is commonly seen in platformer games. In \Cref{fig:parallax} we show an image generated by Imagen2, where taking a latent action moves the foreground at a different rate to the background (as indicated by the length of different colored arrows).

\begin{figure}[h!]
    \centering
    \includegraphics[width=0.99\linewidth]{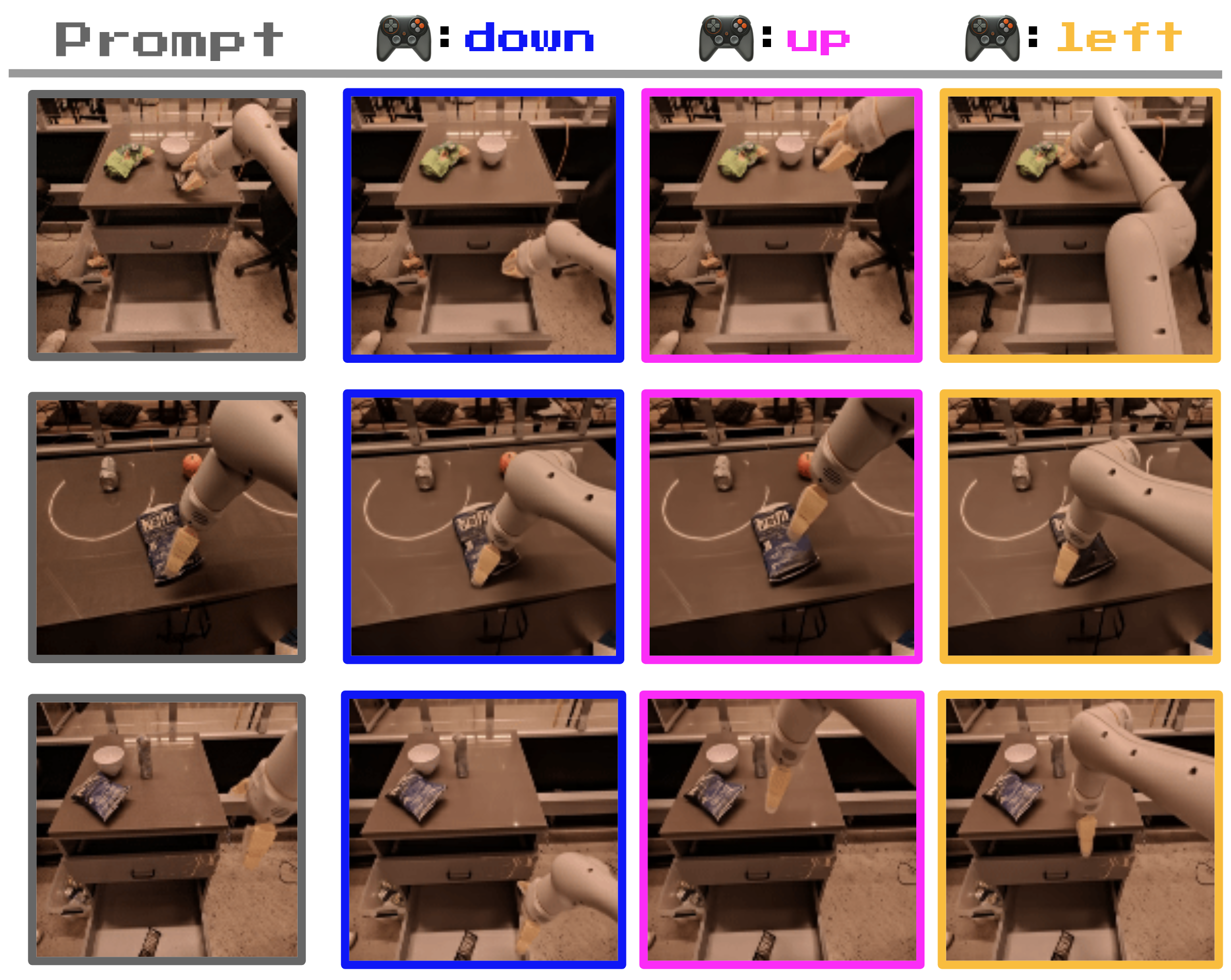}
    \caption{\textbf{Controllable, consistent latent actions in Robotics}: trajectories beginning from three different starting frames from our Robotics dataset. Each column shows the resulting frame from taking the same latent action five times. Despite training without action labels, the same actions are consistent across varied prompt frames and have semantic meaning: \emph{down}, \emph{up} and \emph{left}.}
    \label{fig:action_grid_robotics}
\end{figure}

\textbf{Robotics-trained model} We trained a 2.5B-parameter model on the Robotics dataset using the same hyperparameters found to be best on Platformers, achieving an FVD of 82.7 on the test split. As shown in \Cref{fig:action_grid_robotics}, this model successfully learns distinct and consistent actions from video data, requiring neither text nor action labels (as in e.g. \citet{yang2023learning}). Notably, our model learns not only the controls of the robotic arm but also the interactions and deformations of various objects (\Cref{fig:chips}). We believe this shows our approach presents a path to using larger video datasets from the Internet to create a foundational world model for robotics, with low-level controllable simulation that could be used for a variety of applications.

\subsection{Training Agents}
\begin{figure}[h!]
    \centering
    \includegraphics[width=\linewidth]{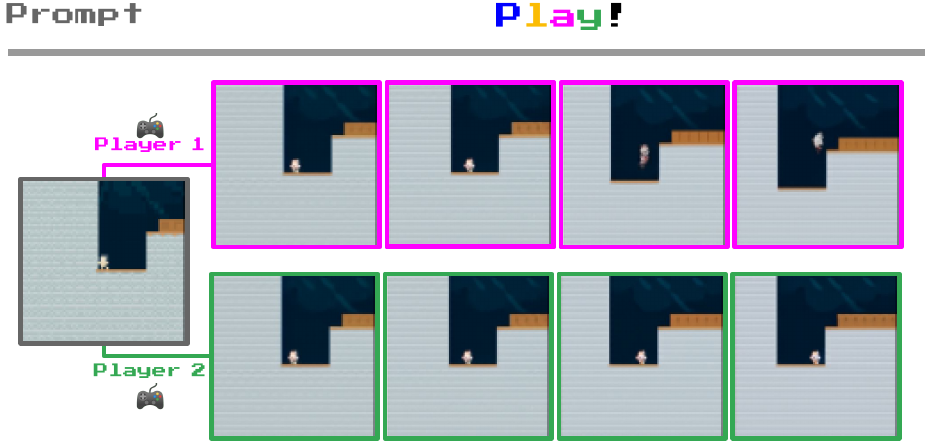}
    \caption{\textbf{Playing from RL environments}: Genie can generate diverse trajectories given an image of an unseen RL environment.}
    \label{fig:coinrun_traj}
\end{figure}

We believe Genie could one day be used as a foundation world model for training generalist agents. In Figure~\ref{fig:coinrun_traj} we show that the model can already be used for generating diverse trajectories in unseen RL environments given starting frames. We further investigate if latent actions learnt from Internet videos can be used for imitating behaviors from unseen videos. We use a frozen LAM to label a sequence of expert videos from a target environment with discrete latent actions and then train a policy that predicts the likelihood of the expert taking a latent action given an observation. We then use a small dataset with expert ground-truth actions for mapping latent to real actions (see \Cref{appendix:bc} for more details).

\begin{figure}[htb]
    \centering
    \vspace{-2.3mm}
    \includegraphics[width=\linewidth]{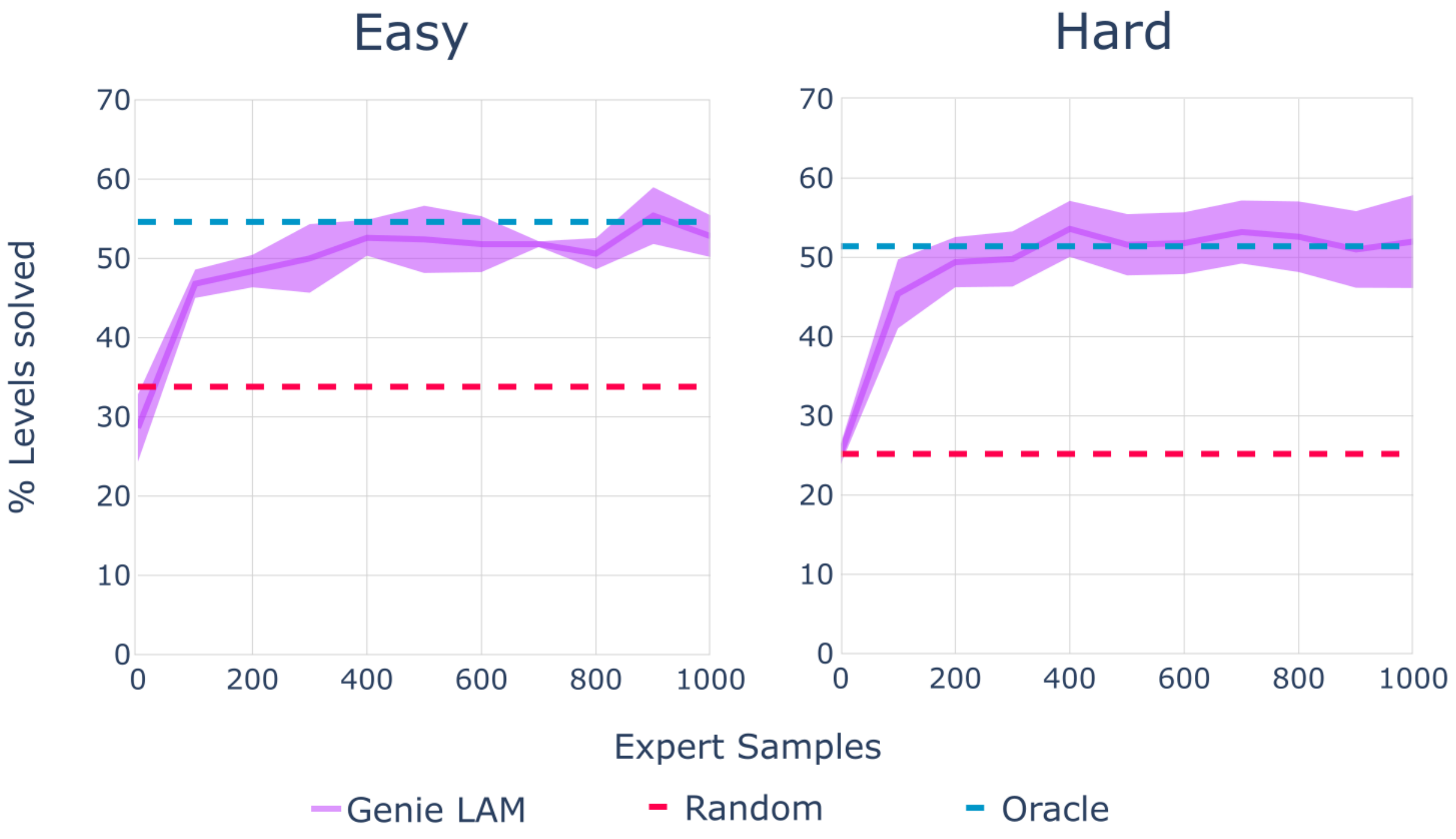}
    \caption{\textbf{BC results}. Mean percentage of levels solved out of 100 samples, averaged over $5$ seeds with $95\%$ confidence intervals.}
    \vspace{-1mm}
    \label{fig:bc}
\end{figure}

We evaluate in both hard and easy settings of a procedurally generated 2D-platformer environment, CoinRun \citep{procgen}, and compare against an oracle behavioral cloning (BC) model that has access to expert actions as an upper bound, and a random agent as a lower bound (Figure~\ref{fig:bc}). The LAM-based policy achieves the same score as the oracle given as few as 200 expert samples to adapt, despite almost certainly never seeing CoinRun before. This provides evidence that the learnt latent actions are consistent and meaningful for transfer, as the mapping from latent to real contains no information about the current observation.
\subsection{Ablation Studies}
\textbf{Design choices for latent action model}
In designing our latent action model, we carefully considered the type of input to use. While we ultimately chose to use the original images (pixels), we evaluated this choice against the alternative of using tokenized images (replacing x with z in \Cref{fig:lam_architecture}). We refer to this alternative approach as the ``token-input" model (see \Cref{tab:ablation_lam}).

While this model achieved a slightly lower FVD score on the Platformers dataset, it did not maintain this advantage on the Robotics dataset. More importantly, in both environments, the token-input model exhibited worse controllability (as measured by \psnrdiff). This suggests that some information about video dynamics and movement might have been lost during tokenization, and as a result it is beneficial for the latent action model to take in raw videos as input.

\begin{table}[h]
\caption{\textbf{Latent action model input ablation}. We see that Genie achieves higher controllability.}
\label{tab:ablation_lam}
\vspace{-3mm}
\centering
\resizebox{.5\textwidth}{!}{
\begin{tabular}{lccccc}
\toprule
& Dataset & \#Params & FVD ($\downarrow$) & \psnrdiff ($\uparrow$) \\ \midrule
Token-input & Platformers & 2.3B & \textbf{38.8}  & 1.33   \\ 
Pixel-input (\textbf{Genie}) & Platformers & 2.5B & 40.1 &  \textbf{1.91} \\ 
\midrule
Token-input & Robotics & 1B & 257.8 & 1.65  \\ 
Pixel-input (\textbf{Genie}) & Robotics & 1B & \textbf{136.4} & \textbf{2.07} \\ 
\bottomrule
\end{tabular}
}
\end{table}

\textbf{Tokenizer architecture ablations} We compare the performance of three choices of tokenizers, including 1) (spatial-only) ViT, 2) (spatial-temporal) ST-ViViT and 3) (spatial-temporal) C-ViViT (\Cref{tab:tokenizer_ablation}). For comparison we use similar number of parameters for all tokenizers, with patch size 10, batch size 128 and sequence length 16. We then train the same dynamics and latent action model on these three different tokenizers, and report their FVD as well as \psnrdiff. 

\begin{table}[h]
\caption{\textbf{Tokenizer architecture ablation}: Our ST-ViViT architecture results in the best performing tokenizer.}
\vspace{-3mm}
\label{tab:tokenizer_ablation}
\centering
\resizebox{.5\textwidth}{!}{
\begin{tabular}{lcccc}
\toprule
& \#Params & Memory & FVD ($\downarrow$) & \psnrdiff ($\uparrow$) \\ \midrule
ViT                                 & 230M & 0.3GB & 114.5 & 1.39 \\ 
C-ViViT \citep{villegas2023phenaki} & 225M & 1.6GB & 272.7 & 1.37 \\ \textbf{ST-ViViT (ours)}            & 205M & 0.9GB & \textbf{81.4} & \textbf{1.66} \\ 
\bottomrule
\end{tabular}
}
\end{table}

Our proposed ST-ViViT architecture provides both improved video generation (FVD) and \psnrdiff, for a reasonable trade-off in memory, as compared to to C-ViViT and the spatial-only ViT. This demonstrates its ability to generate videos of high fidelity and controllability, respectively. While C-ViViT employs a full space-time attention mechanism, resulting in significantly higher memory consumption compared to the other two architectures at the same parameter count, this does not translate to improved performance. In fact, C-ViViT exhibits a tendency towards overfitting, necessitating strong regularization during training, which might explain its considerably lower performance.

\section{4. Related Work}
\textbf{World models}
Generative interactive environments can be considered a class of \emph{World Models} \citep{worldmodels, oh2015}, which enable next-frame prediction that is conditioned on action inputs \citep{actionsurvey, dreamer, hafner2021mastering, micheli2023transformers, robine2023transformerbased, Kim_2020_CVPR, Kim_2021_CVPR, bamfordnge2020, chiappa2017recurrent, NEURIPS2022_9316769a, scienceaar6170}. Such models can be useful for training agents, as they can be used for learning policies without direct environment experience at agent training time. However, learning the models themselves typically requires action-conditioned data obtained directly from the environment. In contrast, our approach seeks to learn a world model in an unsupervised fashion from videos alone. 
Recently, there has been renewed emphasis on scaling world models. GAIA-1 \citep{hu2023gaia1} and UniSim \citep{yang2023learning} learn world models for autonomous driving and robotic manipulation respectively. These approaches require both text and action labels, while we focus on training from video-only data from publicly available Internet videos.

\textbf{Video models} Our work is related to \emph{video models}, which typically condition on initial frames (or text) and predict the remaining frames in a video \citep{vpnkalchbrenner17a, dvdgan, finn2016, trivdgan2020, lotter2017deep, yan2021videogpt, Blattmann2023AlignYL, walker2021predicting, NEURIPS2021_757b505c, hoppe2022diffusion, singer2023makeavideo, ho2022imagen, NEURIPS2022_39235c56, videoworldsimulators2024, 10205485}. Our approach most resembles recent transformer based models such as Phenaki \citep{villegas2023phenaki}, TECO \citep{teco} and MaskViT \citep{gupta2023maskvit}, as we use MaskGIT \citep{Chang_2022_CVPR} and an ST-Transformer \citep{xu2020spatial} over tokenized images. While video models are becoming increasingly controllable (e.g. \citep{layered2022}), we seek a more agentic goal and explicitly learn a \emph{latent action space} from data, allowing users or agents to ``play'' the model using latent action-conditioned predictions.

\textbf{Playable Video Generation} Genie generalizes beyond Playable Video Generation (PVG) \citep{menapace2021}, where latent actions are used for controlling world models learnt directly from videos \citep{menapace2021, Menapace2022PlayableEnvironments}. In contrast to Genie, PVG considers domain-specific static examples, rather than generating entirely new environments via prompting. Thus, scaling beyond this setting required non-trivial architectural changes, dropping inductive biases in exchange for a general method.

\textbf{Environment generation} Our work is also related to \emph{Procedural Content Generation}~\citep[PCG, e.g.][]{pcg, pcg2} where machine learning has proven highly effective for generating game levels \citep{pcgml}, recently via language models that directly write game code \citep{sudhakaran2023prompt, todd2023level}. Language models themselves can also be considered to be interactive environments \citep{wong2023word}, albeit lacking a visual component. By contrast in our setting the levels can be learnt and generated directly from pixels, which enables us to utilize the diversity of Internet video data.

\textbf{Training agents with latent actions} Prior works have used latent actions for imitation from observation \citep{edwards2019imitating}, planning \citep{rybkin2019learning} and pre-training RL agents \citep{ye2022become, schmidt2024learning}. These approaches have similar objectives to our latent action model, though have not been applied at scale. VPT~\citep{baker2022video} is a recent approach that uses an inverse dynamics model learnt from human-provided action labeled data, to label Internet-scale videos with actions that can then be used for training a policy. We showed, in contrast, that we can use \emph{latent} actions learnt from Internet videos to infer policies for arbitrary environments, avoiding the need for ground-truth actions that are costly and may not generalize.

\section{5. Conclusion and Future Work}
We proposed Genie, a new form of generative AI that enables anyone, even children, to dream up, create, and step into generated worlds as we can with human-designed simulated environments. Genie can be prompted to generate a diverse set of interactive and controllable environments despite training from video-only data.

There are clear improvements that can be made to the model. Genie inherits some of the weaknesses of other autoregressive transformer models, and can hallucinate unrealistic futures. And while we have made progress with spatiotemporal representations, we are still limited to 16 frames of memory which makes it challenging to get consistent environments over long horizons. Finally, Genie currently operates around 1FPS and requires future advances to achieve an efficient frame rate for interaction. 

Still, we believe Genie opens up vast potential for future research. Given its generality, the model could be trained from an even larger proportion of Internet videos to simulate diverse, realistic, and imagined environments. Furthermore, we only briefly touched upon the capabilities of using Genie for training agents, but given that the lack of rich and diverse environments is one of the key limitations in RL, we could unlock new paths to creating more generally capable agents. 

\section*{Broader Impact}
\textbf{Societal Impact} Genie could enable a large amount of people to generate their own game-like experiences. This could be positive for those who wish to express their creativity in a new way, for example children who could design and step into their own imagined worlds. We also recognize that with significant advances, it will be critical to explore the possibilities of using this technology to amplify existing human game generation and creativity---and empowering relevant industries to utilize Genie to enable their next generation of playable world development.

\textbf{Training Data and Weights}: We have chosen not to release the trained model checkpoints, the model's training dataset, or examples from that data to accompany this paper or the website.  We would like to have the opportunity to further engage with the research (and video game) community and to ensure that any future such releases are respectful, safe and responsible.

\textbf{Reproducibility}: We understand that it may be challenging for researchers with fewer computational to reproduce our main results. In order to mitigate this issue, we describe a smaller scale, fully reproducible example in \Cref{appendix:case_study} that can run on a single mid-range TPU (or GPU). Given that many design choices translate between the two settings, we believe this will make it possible for the broader community to investigate future architectural improvements as well as additional research directions resulting from our work.

\section*{Acknowledgements}
We thank Mateusz Malinowski, Philip Ball and Louis Kirsch for reviewing a draft of our paper; Cassidy Hardin, David Bridson, Eric Lau, Lars Lowe Sjoesund, Lucas Smaira and Bernardo Avila Pires for help with our Platformers dataset; Ruben Villegas for valuable discussions on our video model training and evaluation; and Adrian Bolton, Rushil Mistry, Hannah Openshaw, Zoubin Ghahramani, Raia Hadsell, Koray Kavukcuoglu, Daan Wierstra, Doina Precup and Ed Hirst for strategic advice and guidance. We make use of the DeepMind Jax ecosystem \citep{babuschkin2010deepmind} and specifically thank Andy Brock for building the internal framework we used for our model training and Arthur Brussee who provided an initial interface that enabled us to ``play'' our models. Finally, thank you to Seneca and Caspian Clune for their creative sketches, potentially making them the youngest ever game designers.

\newpage
\onecolumn
\section*{Author Contributions}
We list authors alphabetically by last name. Please direct all correspondence to Ashley Edwards (\href{mailto:edwardsashley@google.com}{edwardsashley@google.com}) and Jack Parker-Holder (\href{mailto:jparkerholder@google.com}{jparkerholder@google.com}).

\subsection*{Core Contributors}

\begin{itemize}
     \item \textbf{Jake Bruce}: project leadership, video tokenizer research, action model research, dynamics model research, scaling, model demo, infrastructure
     \item \textbf{Michael Dennis}: dynamics model research, scaling, metrics, model demo, infrastructure
     \item \textbf{Ashley Edwards}: genie concept, project leadership, action model research, agent training, model demo
     \item \textbf{Edward Hughes}: dynamics model research, infrastructure
     \item \textbf{Matthew Lai}: dataset curation, infrastructure
     \item \textbf{Aditi Mavalankar}: action model research, metrics, agent training
     \item \textbf{Jack Parker-Holder}: genie concept, project leadership, dynamics model research, scaling, dataset curation
     \item \textbf{Yuge (Jimmy) Shi}: video tokenizer research, dynamics model research, dataset curation, metrics
     \item \textbf{Richie Steigerwald}: dataset curation, metrics
\end{itemize}

\subsection*{Partial Contributors and Advisors}

\begin{itemize}
     \item \textbf{Chris Apps}: project management
     \item \textbf{Yusuf Aytar}: technical advice
     \item \textbf{Sarah Bechtle}: technical advice
     \item \textbf{Feryal Behbahani}: strategic advice
     \item \textbf{Stephanie Chan}: technical advice
     \item \textbf{Jeff Clune}: technical advice, strategic advice
    \item \textbf{Lucy Gonzalez}: project management
    \item \textbf{Nicolas Heess}: strategic advice
    \item \textbf{Simon Osindero}: technical advice
    \item \textbf{Sherjil Ozair}: technical advice
    \item \textbf{Scott Reed}: technical advice
    \item \textbf{Jingwei Zhang}: technical advice
    \item \textbf{Konrad Zolna}: scaling, technical advice
\end{itemize}

\subsection*{Sponsors}
\begin{itemize}
    \item \textbf{Nando de Freitas}: strategic advice
    \item \textbf{Tim Rocktäschel}: genie concept, project leadership
    \item \textbf{Satinder Singh}: strategic advice
\end{itemize}
\newpage
\bibliographystyle{abbrvnat}
\nobibliography*
\bibliography{refs}

\newpage
\appendix
\onecolumn

\section{Additional Example Trajectories}
\label{appendix:more_example}
\begin{figure}[h!]
  \centering 
  \includegraphics[width=0.85\linewidth]{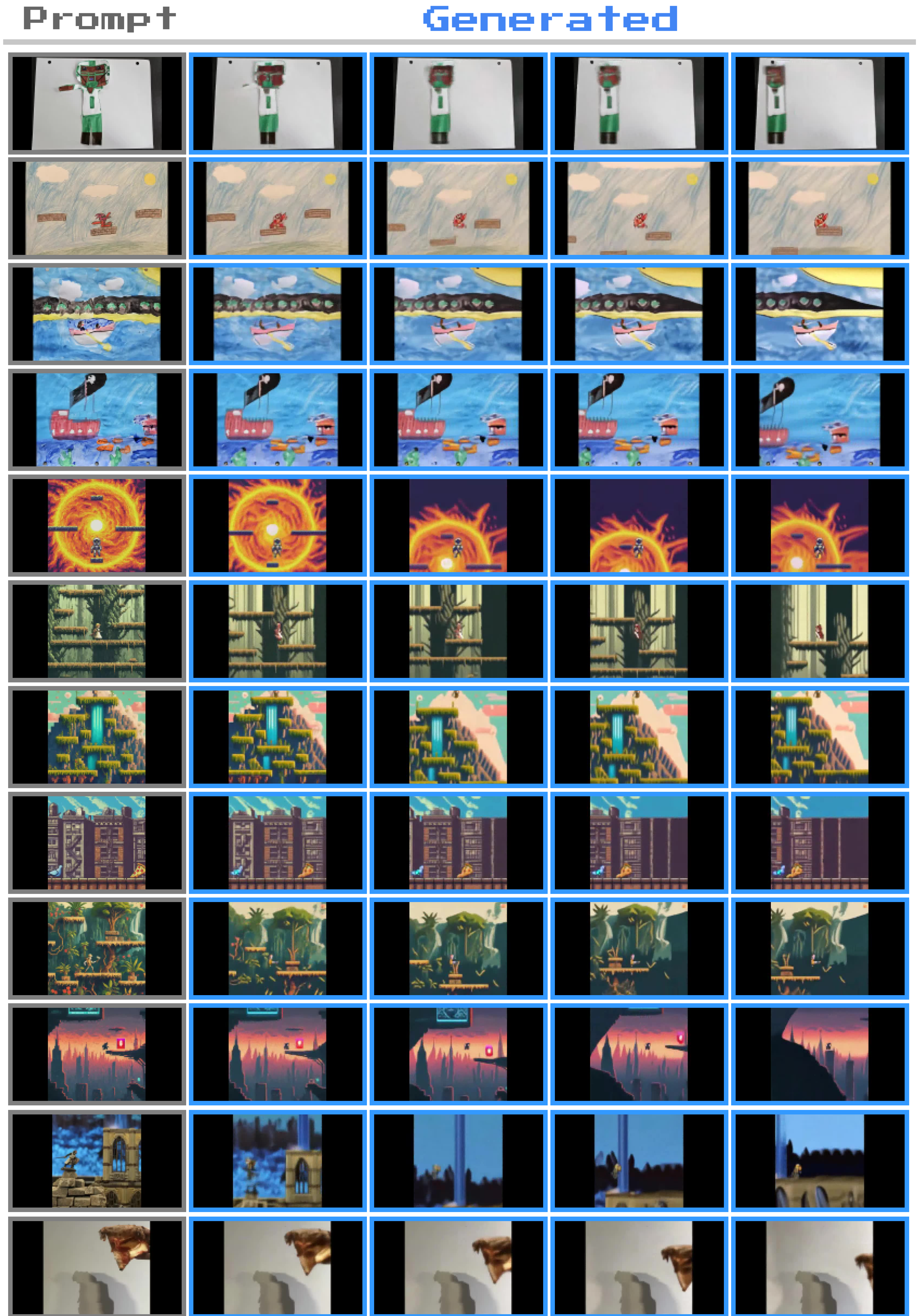}
  \caption{\textbf{More example trajectories}: the model is prompted with either hand-drawn sketches, images generated from text-to-image generative models or realistic photos. Actions that drive the dynamics of the trajectory are provided by human input.}
  \label{fig:more_example}
\end{figure}

\begin{figure}[h!]
    \centering
    \vspace{-2mm}
    \includegraphics[width=0.9\linewidth]{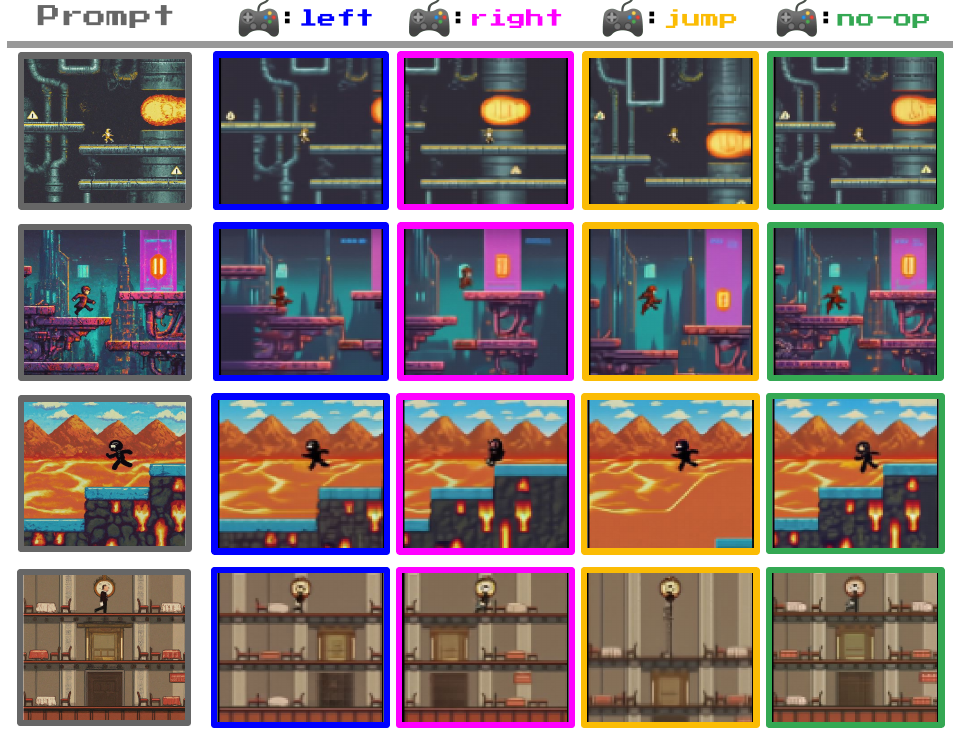}
    \vspace{-1mm}
    \caption{\textbf{Controllable, consistent latent actions in Platformers}: trajectories beginning from four different starting frames from our Platformers dataset. Each column shows the resulting frame from taking the same latent action five times. Despite training without action labels, not only are the same actions consistent across varied prompt frames, but also have semantic meaning: \emph{left}, \emph{right}, \emph{jump}, and \emph{no-op}.}
    \vspace{-3mm}
    \label{fig:consistent_actions_platformers}
\end{figure}

\section{Dataset} \label{sec:app_dataset}

\subsection{Platformers Dataset}
\label{appendix:ytdata}

\paragraph{Initial Dataset} We generated a dataset by filtering publicly available Internet videos, using the following criteria:
\begin{itemize}
    \item The title contains keywords relating to 2D platformer games.
    \item The title or description must contain an action word, such as ``speedrun'' or ``playthrough''.
    \item The title must not contain negating words such as ``movie'' or ``unboxing''.
\end{itemize}
We then split each video into 16s clips at 10 FPS, which corresponds to 160 frames per clip. Our resulting dataset contains 55M videos, which totals around 244k hours. When selecting keywords, we manually spot checked results to check that they typically produced 2D platformer gameplay videos which are not outnumbered by other sorts of videos which happen to share similar keywords.

\paragraph{Filter Pipeline} We noticed that many of the videos in the dataset were of poor quality, impacting our model performance. We propose a scalable approach to systematically filter the data, using a learned classifier as in \cite{baker2022video}. First, we define high quality videos as those that display clear gameplay and do not contain distractor items such as menu screen or streamer faces. We then filter this data as follows:
\begin{enumerate}
    \item Our team hand labelled 10k videos, with roughly ten hours of total human effort. The labels ranged from 5 (best) to 1 (worst) quality.
    \item We trained a 11M parameter ResNet18 \citep{resnet} with binary classification where we deleted all entries rated 2-4 and classified 5 as good and 1 as bad.
    \item We then apply a decision rule based on model prediction and confidence to determine whether to keep the video.
\end{enumerate}
Consistent to findings in prior work \citet{baker2022video, oquab2023dinov2}, having high quality data outweighs the quantity of data -- even though the curated datasaet is only just over 10\% the size of the original dataset, the model trained on the curated dataset outperforms in terms of FVD, see \Cref{tab:dataset_ablation}. Our final dataset is 6.8M videos for a total of over 30k hours.

\begin{table}[h]
\caption{Effect of dataset curation. }
\vspace{3pt}
\label{tab:dataset_ablation}
\centering
\resizebox{.5\textwidth}{!}{
\begin{tabular}{lcc}
\toprule
& \#Params & FVD ($\downarrow$) \\ \midrule
Original dataset (55M videos) & 580M & 61.4 \\ 
Curated dataset (6.8M videos)  & 580M & \textbf{54.8} \\ \bottomrule
\end{tabular}
}
\end{table}

\section{Training details} \label{sec:app_training}
\subsection{Latent Action Model Training}
\label{sec:actionmodel_details}

We found a benefit from increasing the number of codes (i.e. number of actions), at the cost of reduced playability for human and AI agents.

\begin{table}[h]
\caption{Platformers action model hyperparameters}
\label{tab:platformer_actions}
\centering
\small{
\begin{tabular}{llc}
\toprule
Component & Parameter & Value \\ \midrule
Encoder & num\_layers & 20  \\ 
& d\_model  & 1024 \\ 
& num\_heads  & 16 \\ \midrule
Decoder & num\_layers & 20  \\ 
& d\_model  & 1024 \\ 
& num\_heads  & 16 \\ \midrule
Codebook & num\_codes & 8  \\ 
 & patch\_size & 16  \\ 
& latent\_dim  & 32 \\ \bottomrule
\end{tabular}}
\end{table}

Note that the model inputs are normalized between $0$ and $1$ and the final outputs of the decoder are placed through a sigmoid. 

\subsection{Video Tokenizer Training}
\label{sec:vidtok_details}

Here we describe our video tokenizer training. We found it more effective to scale our decoder than the encoder, and a marginal gain from increasing batch size (see Table~\ref{tab:batch_size_hps_tokenizer}).

\begin{table}[h]
\centering
\caption{Tokenizer batch size scaling hyperparameters.}
\label{tab:batch_size_hps_tokenizer}
\resizebox{.5\textwidth}{!}{
\begin{tabular}{lccc}
\toprule
batch\_size & training hardware & FLOPs & PSNR \\ \midrule
64 & 64 TPUv2 & $4.22 \times 10^{20} $ & 35.7 \\ 
384 & 64 TPUv3 & $2.57 \times 10^{21} $ & 36.5 \\ 
\bottomrule
\end{tabular}
}
\end{table}

\begin{table}[h]
\caption{Platformers video tokenizer hyperparameters.}
\label{tab:platformers_vid_tok}
\centering
\small{
\begin{tabular}{llc}
\toprule
Component & Parameter & Value \\ \midrule
Encoder & num\_layers & 12  \\ 
& d\_model  & 512 \\ 
& num\_heads  & 8 \\ 
& k/q\_size  & 64 \\ \midrule
Decoder & num\_layers & 20  \\ 
& d\_model  & 1024 \\ 
& num\_heads  & 16 \\
& k/q\_size  & 64 \\ \midrule
Codebook & num\_codes & 1024  \\ 
  & patch\_size & 4  \\ 
& latent\_dim  & 32 \\ \bottomrule
\end{tabular}}
\end{table}

We train our video tokenizer for 300k steps using the AdamW optimizer, with cosine decay, using the hyperparameters in Table~\ref{tab:vid_tok_adam}.

\begin{table}[h]
\caption{Video tokenizer optimizer hyperparameters}
\label{tab:vid_tok_adam}
\centering
\small{
\begin{tabular}{lc}
\toprule
Parameter & Value \\ \midrule
max\_lr & 3e-4  \\ 
min\_lr  & 3e-4 \\ 
$\beta_1$  & 0.9 \\ 
$\beta_2$  & 0.9 \\ 
weight\_decay & 1e-4 \\ 
warmup\_steps & 10k \\ \bottomrule
\end{tabular}}
\end{table}

\subsection{Dynamics Model Training}
\label{sec:dynamicsmodel_details}

\begin{table}[h]
\caption{Dynamics model optimizer hyperparameters}
\label{tab:dynamics_model_adam}
\centering
\small{
\begin{tabular}{lc}
\toprule
Parameter & Value \\ \midrule
max\_lr & 3e-5  \\ 
min\_lr  & 3e-6 \\ 
$\beta_1$  & 0.9 \\ 
$\beta_2$  & 0.9 \\ 
weight\_decay & 1e-4 \\ 
warmup\_steps & 5k \\ \bottomrule
\end{tabular}}
\end{table}

\section{Scaling Experiments Details} \label{sec:app_scaling_details}

In this section we provide more details on the architecture as well as compute budget for the scaling experiments. 

\paragraph{Scaling model size} For all models we use a batch size of 256. We train all models for 200k steps, thus use a total of 750B training tokens for each run. All runs make use of batch parallelism and stage-3 ZeRO sharding \citep{rajbhandari2020zero}, while our larger models also make use of tensor parallelism \citep{megatron}. For this experiment we make use of TPUv2 and TPUv3 \citep{jouppi2020domain}. See Table~\ref{tab:model_size_hps} for more details.

\begin{table}[h]
\centering
\caption{Model size scaling architectures and compute usage. All models were trained for 200k steps with a batch size of 256, equating to 750B tokens.}
\label{tab:model_size_hps}
\resizebox{.85\textwidth}{!}{
\begin{tabular}{lccccccc}
\toprule
Parameters & num\_layers & num\_heads & d\_model & k/q size & training hardware & training time & FLOPs \\ \midrule
41M & 18 & 8 & 512 & 64 & 64 TPUv2 & 3 days & $2.05 \times 10^{20}$ \\ 
96M & 16 & 16 & 768 & 64 & 64 TPUv2 & 6 days & $3.58 \times 10^{20}$ \\ 
192M & 20 & 18 & 1024 & 64 & 64 TPUv2 & 9 days & $6.4 \times 10^{20}$ \\ 
404M & 21 & 12 & 1536 & 128 & 64 TPUv2 & 18 days & $1.2 \times 10^{21}$ \\
811M & 20 & 20 & 2048 & 128 & 128 TPUv3 & 7 days & $2.2 \times 10^{21}$ \\
1.6B & 28 & 22 & 2560 & 128 & 128 TPUv3 & 12 days & $4.04 \times 10^{21}$ \\
2.7B & 36 & 22 & 3072 & 128 & 256 TPUv3 & 16 days & $6.91 \times 10^{21}$ \\ 
\bottomrule
\end{tabular}
}
\end{table}

\paragraph{Scaling batch size}  All models use the same architecture with 2.3B parameters, as shown in Table~\ref{tab:batch_size_hps}, and train for 200k steps. The only difference between the three runs is hardware---the 128, 256 and 448 batch size models train on 64 TPUv3, 128 TPUv3 and 64 TPUv5p respectively.  

\begin{table}[h]
\centering
\caption{Batch size scaling hyperparameters. All models use the following architecture for 200k steps, differing only in batch size.}
\label{tab:batch_size_hps}
\resizebox{.5\textwidth}{!}{
\begin{tabular}{lcccc}
\toprule
Parameters & num\_layers & num\_heads & d\_model & k/q size \\ \midrule
2.3B & 34 & 20 & 2560 & 128 \\ 
\bottomrule
\end{tabular}
}
\end{table}

\paragraph{Genie Model} The parameter count, model architecture as well as compute usage of the dynamics model for the final Genie model is listed in \Cref{tab:genie_11b}. We train a 10.1B dynamics model with a batch size of 512, for a total of 125k steps using 256 TPUv5. 
\begin{table}[h]
\centering
\caption{Genie dynamics model hyperparameters.}
\label{tab:genie_11b}
\resizebox{.55\textwidth}{!}{
\begin{tabular}{lccccc}
\toprule
Parameters & num\_layers & num\_heads & d\_model & k/q size & FLOPs \\ \midrule
10.1B & 48 & 36 & 5120 & 128 & $6.6\times 10^{22}$ \\ 
\bottomrule
\end{tabular}
}
\end{table}

\section{Behavioral Cloning Details}
\label{appendix:bc}
In this section we provide more details about our behavioral cloning experiments. We train within the Procgen CoinRun environment~\citep{procgen} and evaluate in a held out test set. We assume we have a dataset of expert sequences in this environment from an agent trained with R2D2~\citep{kapturowski2018recurrent}. We then train an agent to imitate from this data. Notably, the oracle agent has access to the corresponding ground-truth expert actions. We now discuss how we can utilize a pre-trained LAM to infer the actions taken.

\subsection{Genie LAM}
In order to train an agent to imitate from unseen videos, we can use a frozen LAM from a Genie model trained on Internet videos. Given an expert sequence $\langle x_t, x_{t+1} \rangle$ we extract the corresponding latent action label $a_{t} \leftarrow LAM(x_t, x_{t+1})$. We then train a policy $\pi(a_t|x_t)$ to predict the likelihood of the expert taking latent action $a_t$ given observation $x_t$. Note that this procedure is similar to prior works that learn from videos \citep{torabi2018behavioral, baker2022video}. However, these approaches use ground-truth actions for labeling videos whereas we utilize latent actions learnt completely offline.

During inference, we must map latent actions emitted by the policy to real actions. To do this, we utilize a small set of action-labeled expert sequences. Given an expert sequence $\langle x_t, u_t, x_{t+1} \rangle$ (we denote $u_t$ for ground-truth actions to avoid confusion with predicted latent actions), we use the LAM to obtain a latent action $a_t$ and fill a dictionary $D$ consisting of mapped latents to a list of corresponding real actions. In summary, given an observation $x_t$ from the environment, we can obtain the most likely latent action as $a_t \sim \pi(s_t)$, and then take the corresponding real action as $u_t \sim D[a_t]$.

Note that other works have used data extracted from the agent's policy to obtain a mapping from latent to real actions \citep{edwards2019imitating, ye2022become}, but we found using expert data enabled us to better evaluate the quality of the learnt policy. As shown in the main text, the agent was capable of adapting with as few as $200$ expert labels. 

\subsection{Architecture}
We train a transformer as the policy for both the oracle and latent BC agents. We utilize our proposed ST-ViViT architecture for encoding the frames $\x_{1:t}=(x_1,\cdots x_t)$ . All previous actions are placed through a one-hot and then combined with the corresponding frame encoding as an additive embedding. We use a sequence length of $4$ during both training and inference and a batch size of $16$. 

\begin{table}[h!]
\caption{BC model optimizer hyperparameters}
\label{tab:bc_model_adam}
\centering
\small{
\begin{tabular}{lc}
\toprule
Parameter & Value \\ \midrule
max\_lr & 3e-5  \\ 
min\_lr  & 3e-6 \\ 
$\beta_1$  & 0.9 \\ 
$\beta_2$  & 0.96 \\ 
weight\_decay & 1e-4 \\ 
warmup\_steps & 5k \\ \bottomrule
\end{tabular}}
\end{table}

\begin{table}[h!]
\caption{BC policy hyperparameters}
\label{tab:bc_actions}
\centering
\small{
\begin{tabular}{llc}
\toprule
Component & Parameter & Value \\ \midrule
Encoder & num\_layers & 12  \\ 
& d\_model  & 512 \\ 
& patch\_size  & 4 \\ \midrule
Policy & linear\_layer & 512  \\ \bottomrule
\end{tabular}}
\end{table}

Both the oracle and Genie LAM are trained with a cross-entropy loss where targets are either real or latent actions, respectively. During inference, we obtain the final prediction by sampling from the predicted logits. Note we found the oracle agent performed better when we randomly sampled actions $10\%$ of the time.

\section{Reproducible Case Study} \label{appendix:case_study}

In this section we describe a self-contained, fully reproducible case study that can be trained with a single mid range TPU/GPU in under a week. 

\subsection{Data Collection}

First we need to collect the data to train our model. We use the CoinRun environment from the Procgen benchmark \citep{procgen} since it has thousands of visually diverse levels with fairly simple platformer-like dynamics. Using the ``hard'' mode, we collect data using a random policy with no action repeats. We sample level seeds between zero and 10,000 and collect 1,000 timesteps for each level, for a total of 10M transitions.

\subsection{Video Tokenizer Training}
 
Our video tokenizer for CoinRun follows the same setup as described in Section~\ref{sec:model_components}, trained with the optimizer configuration as in Section~\ref{sec:vidtok_details}. The primary difference in this example is we use smaller model sizes (see Table~\ref{tab:coinrun_vid_tok}), and then use a batch size of 48 sequences, of length 16, for a total of 768 images per batch. This is sufficient to fit in a single TPU with 16G memory. The model is trained for three days using a single TPU which is sufficient to complete 300k steps.  

\begin{table}[h]
\caption{CoinRun video tokenizer hyperparameters}
\label{tab:coinrun_vid_tok}
\centering
\small{
\begin{tabular}{llc}
\toprule
Component & Parameter & Value \\ \midrule
Encoder & num\_layers & 8  \\ 
& d\_model  & 512 \\ 
& num\_heads  & 8 \\ \midrule
Decoder & num\_layers & 8  \\ 
& d\_model  & 512 \\ 
& num\_heads  & 8 \\ \midrule
Codebook & num\_codes & 1024  \\ 
  & patch\_size & 4  \\ 
& latent\_dim  & 32 \\ \bottomrule
\end{tabular}}
\end{table}

\subsection{Dynamics + Latent Action Model Training}

Once we have trained the video tokenizer we can then jointly train the latent action and dynamics models. Once again we seek to fit our model training inside 16G memory, so we use a batch size of 36 sequences consisting of 16 frames each, for a total of 576 images. We train both the latent action model and dynamics model in parallel, using the setup described above (see: Section~\ref{sec:actionmodel_details} for the latent action model and Section~\ref{sec:dynamicsmodel_details} for the dynamics model).

We train both the latent action and dynamics models in parallel for 200k steps, using the optimizer hyperparameters in \Cref{tab:dynamics_model_adam}. We find this model generates consistent playable latent actions, resembling the original environment.

\begin{table}[h]
\caption{CoinRun action model hyperparameters}
\label{tab:coinrun_actions}
\centering
\small{
\begin{tabular}{llc}
\toprule
Component & Parameter & Value \\ \midrule
Encoder & num\_layers & 8  \\ 
& d\_model  & 512 \\ 
& num\_heads  & 8 \\ \midrule
Decoder & num\_layers & 8  \\ 
& d\_model  & 512 \\ 
& num\_heads  & 8 \\ \midrule
Codebook & num\_codes & 6  \\ 
& latent\_dim  & 32 \\ \bottomrule
\end{tabular}}
\end{table}

\begin{table}[h]
\caption{CoinRun dynamics model hyperparameters}
\label{tab:coinrun_dynamics}
\centering
\small{
\begin{tabular}{lcc}
\toprule
Component & Parameter & Value \\ \midrule
Architecture & num\_layers & 12  \\ 
& d\_model  & 512 \\ 
& num\_layers  & 8 \\ 
Sampling & temperature  & 1.0 \\ 
& maskgit\_steps  & 25 \\ \bottomrule
\end{tabular}}
\end{table}

\end{document}